\documentclass[11pt,a4paper]{article}
\pdfoutput=1
\usepackage[hyperref]{naaclhlt2018}
\usepackage{times}
\usepackage{latexsym}
\usepackage{algpseudocode}
\usepackage{algorithm}
\usepackage{amsmath}
\usepackage{dblfloatfix}
\usepackage{url}
\usepackage{booktabs}
\usepackage{multirow}
\usepackage{tabularx}

\aclfinalcopy % Uncomment this line for all SemEval submissions

\title{Zoho at SemEval-2019 Task 9: Semi-supervised Domain Adaptation using Tri-training for Suggestion Mining}
\author{ Sai Prasanna \\
  Zoho \\
  {\tt \small saiprasanna.r@zohocorp.com} \\
  {\tt \small sai.r.prasanna@gmail.com} \\\And
  Sri Ananda Seelan \\
  Zoho \\
  {\tt \small anandaseelan.ln@zohocorp.com} \\}

\date{}

\begin{document}
\maketitle
\begin{abstract}
  This paper describes our submission for the SemEval-2019 Suggestion Mining task. A simple Convolutional Neural Network (CNN) classifier with contextual word representations from a pre-trained language model is used for sentence classification. The model is trained using tri-training, a semi-supervised bootstrapping mechanism for labelling unseen data. Tri-training proved to be an effective technique to accommodate domain shift for cross-domain suggestion mining (Subtask B) where there is no hand labelled training data. For in-domain evaluation (Subtask A), we use the same technique to augment the training set. Our system ranks thirteenth in Subtask A with an $F_1$-score of 68.07 and third in Subtask B with an $F_1$-score of 81.94.
\end{abstract}

\section{Introduction}

Task 9 of SemEval-2019 \cite{negi2019semeval} focuses on mining sentences that contain suggestions in online discussions and reviews. Suggestion Mining is modelled as a sentence classification task with two Subtasks:
\begin{itemize}
    \item{Subtask A evaluates the classifier performance on a technical domain specific setting.}
    \item{Subtask B evaluates the domain adaptability of a model by doing cross-domain suggestion classification on hotel reviews.}
\end{itemize}

We approached this task as an opportunity to test the effectiveness of transfer learning and semi-supervised learning techniques. In Subtask A, the high class imbalance and relatively smaller size of the training data made it an ideal setup for evaluating the efficacy of recent transfer learning techniques. Using pre-trained language models for contextual word representations has been shown to improve many Natural Language Processing (NLP) tasks \cite{Peters2018DeepCW, Ruder2018UniversalLM, Radford2018ImprovingLU, Devlin2018BERTPO}. This transfer learning technique is also an effective method when less labelled data is available as shown in \cite{Ruder2018UniversalLM}. In this work, we use the BERT model \cite{Devlin2018BERTPO} for obtaining contextual representations. This results in enhanced scores even for simple baseline classifiers.

Subtask B requires the system to not use manually labelled data and hence it lends itself to a classic semi-supervised learning scenario. Many methods have been proposed for domain adaptation for NLP \cite{Blitzer2007BiographiesBB, Chen2011CoTrainingFD, Chen2018MultinomialAN, Zhou2005TritrainingEU, Blum1998CombiningLA}. We use a label bootstrapping technique called tri-training \cite{Zhou2005TritrainingEU} with which unlabelled samples are labelled iteratively with increasing confidence at each training iteration(explained in Section \ref{tri}). \citet{Ruder2018StrongBF} shows the effectiveness of tri-training for baseline deep neural models in text classification under domain shift. They also propose a multi-task approach for tri-training, however we only adapt the classic tri-training procedure presented for suggestion mining task.

Detailed explanation of the submitted system and experiments are elicited in the following sections. Section \ref{systemdesc} describes the components of the system. Following this, Section \ref{experiments} details the experiments, results and ablation studies that were performed.

\section{System Description} \label{systemdesc}
The models and the training procedures are built using AllenNLP library \cite{Gardner2018AllenNLPAD}. All the code to replicate our experiments are public and can be accessed from \url{https://github.com/sai-prasanna/suggestion-mining-semeval19}.

\subsection{Data cleaning and pre-processing}

Basic data pre-processing is done to normalize whitespace, remove noisy symbols and accents. Very short sentences with less than four words are disregarded from training.

\subsection{Word Representations}
We use GloVe word representations \cite{Pennington2014GloveGV} and compare the performance improvement that we obtain with pre-trained BERT representations \cite{Devlin2018BERTPO}.

\subsection{Suggestion Classification} \label{classificationmodel}
Our baseline classifier is Deep Averaging Network (DAN) \cite{Iyyer2015DeepUC}. DAN is a neural bag-of-words model that is considered as a strong baseline for text classification. In DAN, a sentence representation is obtained by averaging the word level representations and is fed to a series of rectified linear unit (ReLU) layers with a final softmax layer.

A simple Convolutional Neural Network (CNN) text classifier \cite{Kim2014ConvolutionalNN} is used for the final submission.

\subsection{Training} \label{tri}
We use the classic tri-training procedure for label bootstrapping as mentioned in \cite{Ruder2018StrongBF}. Consider a labelled dataset $L$ from the source domain $S$ and an unlabelled dataset $U$ from the target domain $T$. The objective of tri-training is to label $U$ iteratively and augment it with $L$. Three CNN +BERT classifiers $M_1$, $M_2$, $M_3$ are trained separately using subsets of $L$ namely $l_1$, $l_2$, $l_3$ respectively. These subsets are obtained from $L$ using bootstrap sampling with replacement.

The above mentioned models are used to predict labels for the unlabelled set $U$. Predictions which are agreed by two models is considered as a new training example for the third model in the next iteration. For example, an unlabelled sentence $U_1 \in U$ is added as a labelled example to $l_1$, if and only if the label for $U_1$ is agreed upon by both $M_2$ and $M_3$. Same way, $l_2$ is updated with newly labelled data if those labels have been agreed by $M_1$ and $M_3$ and so on. This constitutes a single iteration of tri-training. The procedure that is used for the training of our models is mentioned in Algorithm \ref{alg:trialgo}.
 
In this way, the original training data gets added with three newly labelled subsets which are again used for the next training iteration. At the end of each iteration, validation $F_1$-score is calculated by using the predictions that are obtained through a majority vote. The procedure is continued until there is no improvement in the validation score.

\begin{algorithm} 
\caption{Tri-training}
\begin{algorithmic}[1]
\State $L \gets Labelled \; Data \; , \; |L| \; = \; m$
\State $U \gets Unlabelled \; Data \; , \; |U| \; = \; n$
\For{$i \gets 1,2,3$}
    \State $l_i \gets BootstrapSamples(L)$
\EndFor
\Repeat
    \For{$i \gets 1,2,3$}
        \State $M_i \gets Train(l_{i})$
    \EndFor
    \For{$i \gets 1,2,3,$}
        \State $l_{i} \gets L$
        \For{$j \gets 1,n$}
            \If{$M_p(U_j) == M_q(U_j) \, $ \\ \hspace{2.5cm} 
                $\, where \, \, p, \, q  \, \neq \, i$}
                \State $l_{i} \gets l_{i} + \{(U_j, M_p(U_j)\}$
            \EndIf
        \EndFor
    \EndFor
\Until{no improvement in validation metrics}
\end{algorithmic}
\label{alg:trialgo}
\end{algorithm}

\section{Experiments and Results} \label{experiments}

\begin{table*}[t]
  \small
  \begin{tabularx}{\textwidth}{Xcccccc}
    \toprule
    \multicolumn{7}{c}{\textbf{Subtask A - Technical Domain}} \vspace{1ex} \\
    \multirow{2}{*}{\textbf{Experiment}} &
      \multicolumn{3}{c}{\textbf{Validation}} &
      \multicolumn{3}{c}{\textbf{Test}} \vspace{1ex} \\ 
        &
      \multicolumn{1}{c}{\textbf{Precision}} &
      \multicolumn{1}{c}{\textbf{Recall}} & 
      \multicolumn{1}{c}{$\mathbf{F_{1}}$} &
      \multicolumn{1}{c}{\textbf{Precision}} &
      \multicolumn{1}{c}{\textbf{Recall}} & 
      \multicolumn{1}{c}{$\mathbf{F_{1}}$} 
       \\
      \midrule
    Organizer Baseline&58.72 & 93.24 & 72.06 & 15.69 & 91.95 & 26.80 \\
    DAN +glove&68.51$\pm$2.43 & 87.30$\pm$5.00 & 76.69$\pm$1.06 & 25.40$\pm$3.56 & 84.60$\pm$9.87 & 38.84$\pm$3.10\\
    DAN +bert&76.06$\pm$1.31 & 90.27$\pm$1.71 & 82.55$\pm$0.50 & 45.80$\pm$4.49 & 90.80$\pm$1.75 & 60.82$\pm$3.99\\
    DAN +bert w/o upsampling&79.04$\pm$2.67 & 83.38$\pm$2.73 & 81.11$\pm$0.68 & 55.06$\pm$6.36 & 83.68$\pm$2.75 & 66.28$\pm$4.28\\
    CNN +bert&80.34$\pm$4.21 & 89.93$\pm$4.23 & 84.76$\pm$0.52 & 50.34$\pm$6.70 & 91.72$\pm$2.55 & 64.81$\pm$4.86\\
    CNN +bert w/o upsampling&83.22$\pm$3.01 & 84.73$\pm$3.86 & 83.90$\pm$0.70 & 58.98$\pm$5.41 & 88.05$\pm$1.63 & \textbf{70.58}$\pm$4.24\\
    CNN +bert +tritrain$_{Test}$\textbf{*}&83.06$\pm$1.96 & 89.19$\pm$1.88 & \textbf{86.00}$\pm$0.35 & 52.89$\pm$2.69 & 90.80$\pm$2.02 & 66.81$\pm$1.90\\
    \midrule
    \multicolumn{7}{c}{\textbf{Subtask B - Hotel Reviews Domain}} \vspace{1ex} \\
    \multirow{2}{*}{\textbf{Experiment}} &
      \multicolumn{3}{c}{\textbf{Validation}} &
      \multicolumn{3}{c}{\textbf{Test}} \vspace{1ex} \\ 
        &
      \multicolumn{1}{c}{\textbf{Precision}} &
      \multicolumn{1}{c}{\textbf{Recall}} & 
      \multicolumn{1}{c}{$\mathbf{F_{1}}$} &
      \multicolumn{1}{c}{\textbf{Precision}} &
      \multicolumn{1}{c}{\textbf{Recall}} & 
      \multicolumn{1}{c}{$\mathbf{F_{1}}$} 
       \\
      \midrule
    Organizer Baseline&72.84 & 81.68 & 77.01 & 68.86 & 78.16 & 73.21 \\
    DAN +glove&82.00$\pm$4.25 & 52.97$\pm$9.25 & 64.01$\pm$5.75 & 73.32$\pm$3.50 & 46.09$\pm$7.21 & 56.35$\pm$4.71\\
    DAN +bert&89.75$\pm$2.79 & 65.74$\pm$8.71 & 75.65$\pm$5.10 & 78.90$\pm$4.03 & 64.20$\pm$8.77 & 70.49$\pm$4.09\\
    DAN +bert w/o upsampling&94.26$\pm$1.87 & 31.73$\pm$5.73 & 47.31$\pm$6.27 & 87.98$\pm$3.41 & 31.09$\pm$7.17 & 45.62$\pm$7.47\\
    CNN +bert&93.77$\pm$1.34 & 51.88$\pm$6.88 & 66.65$\pm$5.68 & 90.17$\pm$2.45 & 50.34$\pm$8.71 & 64.31$\pm$6.72\\
    CNN +bert w/o upsampling&93.94$\pm$1.36 & 45.99$\pm$7.59 & 61.53$\pm$6.73 & 89.75$\pm$4.41 & 44.08$\pm$9.38 & 58.66$\pm$7.79\\
    CNN +bert +tritrain$_{Test}$\textbf{*}&91.91$\pm$2.06 & 88.32$\pm$2.05 & \textbf{90.05}$\pm$0.76 & 81.26$\pm$1.63 & 83.16$\pm$1.40 & \textbf{82.19}$\pm$1.03\\
    CNN +bert +tritrain$_{Yelp}$&88.09$\pm$0.62&87.13$\pm$0.38&87.61$\pm$0.42&78.01$\pm$5.42&86.67$\pm$3.96&81.98$\pm$2.05\\
    \bottomrule \\
  \end{tabularx}
  \caption{Performance metrics of different models on validation and test sets of both subtasks. Confidence intervals for the metrics  are reported for five runs using different random seeds on t-distribution with 95\% confidence. Upsampling is used in the training dataset unless otherwise specified. Single model from experiments with \textbf{*} was used for the final submission.}
  
  \label{table:1}
\end{table*}

This section details the various experiments that were performed using the above components for our submissions.

\subsection{Data}

The test set provided during the trial phase of the evaluation is used as the validation data for all our experiments. For those experiments that do not involve tri-training, we only use the provided labelled data from the technical domain for training.

In Subtask B, for those experiments that involve tri-training, $L$ is the same as mentioned above. $U$ here is obtained in two ways:
\begin{itemize}
    \item Unlabelled data from final test set of Subtask B.
    \item Unlabelled data from Yelp hotel reviews \cite{Blomo2013RecSysC2}.
\end{itemize}

The results reported are mean and confidence intervals of Precision, Recall and $F_1$-score over five runs of the same experiments with different random seeds.

\subsection{Input}

For input representations, we use 300d GloVe vectors with dropout \cite{Srivastava2014DropoutAS} of 0.2 for regularization. We also experiment with the pre-trained BERT base uncased model. The BERT model is not fine-tuned during our training. A dropout of 0.5 is applied for the 768d representations obtained from BERT.

\subsection{Baseline Deep Averaging Net} 

Our neural baseline is Deep Averaging Net (DAN) (Section \ref{classificationmodel}). When used with GloVe, the hidden sizes of DAN are 300, 150, 75, and 2 respectively. When BERT representations are used, the hidden sizes of the network are 768, 324, 162, and 2 respectively. We report an $F_1$-score of 60.82 when DAN is used with BERT in Subtask A and 70.49 in Subtask B. Both these scores are a significant improvements from those obtained with GloVe representations (Table \ref{table:1}).

We retain the same configuration of BERT embedding layer for other experiments also. Training is performed with Adam  \cite{Kingma2015AdamAM} optimizer with a learning rate of $1\mathrm{e}{-3}$ for all the models.

\subsection{CNN Classifier}

 The CNN classifier is composed of four 1-D convolution layers with filter widths ranging from two to five. Each convolutional layer has 192 filters. The output from each layer is max-pooled over sequence (time) dimension. This results in four 192d vectors, which are concatenated to get a 768d output.

The max-pooled outputs are passed through four fully connected feed forward layers with hidden dimensions of 768, 324, 162, and 2 respectively. The intermediate layers use ReLu activation and the final layer is a softmax layer. We use dropout of 0.2 on all layers of the feed forward network except for the final layer.

Without tri-training, this model obtains an absolute improvement of $\approx$ 4\% $F_1$-score over DAN in Subtask A. However in Subtask B, it performs poorer than the baseline DAN model with an $F_1$-score of 64.31. This decrease in performance could be because of overfitting on the source domain due to the larger number of parameters in CNN compared to DAN.

\subsection{Tri-training}
The aim of doing tri-training is for domain adaptation by labelling unseen data from a newer domain. For Subtask B, the CNN + BERT model achieves an $F_1$-score of 82.19 when trained with the tri-training procedure mentioned in Algorithm \ref{alg:trialgo}. Tri-training is used to label the 824 unlabelled sentences from the test set of Subtask B and augmented with the original training data. This score is a huge improvement from the classifier model trained only on the given data which gets an $F_1$-score of 64.31.

We also do the same experiment using 5000 unlabelled sentences from Yelp hotel reviews dataset \cite{Blomo2013RecSysC2}. The model obtains a similar score of 81.98 which proves the importance of tri-training in domain adaptation.

For Subtask A, we get an improvement in the $F_1$-score using tri-training, however the increase is not as profound as we observe for Subtask B. We compare the statistical significance of the different models and experiments in Section \ref{stat}.

\subsection{Upsampling}

We also wanted to find how the class balance in the dataset has affected our model performance. The class distribution of the datasets including the test set distribution that was obtained after the final evaluation phase are mentioned in Table \ref{table:2}. 

\begin{table}[h]
  \centering
  \begin{tabular}{@{}lc@{}}
    \toprule
    Dataset    & Suggestions (\%) \\
    \midrule
    Training              & 23         \\ 
    Subtask A validation  & 50         \\ 
    Subtask B validation  & 50         \\
    Subtask A Test        & \textbf{10}       \\
    Subtask B Test        & 42         \\
    \bottomrule
  \end{tabular}
  \caption{Label distribution}
  \label{table:2}
\end{table}
\begin{table*}[!t]
  \centering
  \begin{tabular}{@{}clll@{}}
    \toprule
    Subtask & Model A                        & Model B                                      &   p-value              \\
    \midrule
    A       & DAN +glove                     & DAN +bert                                    &   $\approx 0$                    \\
    A       & DAN +bert                      & CNN +bert                                    &   0.046                \\ 
    A       & CNN +bert                      & CNN +bert +tritrain$_{Test}$                  &   \textbf{0.108}       \\
    B       & DAN +glove                     & DAN +bert                                    &   $1.419e-05$            \\
    B       & DAN +bert                      & CNN +bert                                    &   \textbf{0.4208}               \\ 
    B       & CNN +bert                      & CNN +bert +tritrain$_{Test}$                   &   $3.251e-08$             \\
    B       & CNN +bert +tritrain$_{Test}$     & CNN +bert +tritrain$_{Yelp}$               &   \textbf{0.5862}       \\
    \bottomrule
  \end{tabular}
  \caption{Pairwise comparison of various models using the McNemar's Test. $p  \leq  0.05$ indicates a significant disagreement between the model predictions.}
  \label{table:3}
\end{table*}

The original training data has a class imbalance with only 23\% of the sentences labelled as suggestions. We tried to balance the labels by naive upsampling, ie., adding duplicates of sentences that are labelled as suggestions. This allowed us to have a balanced training dataset for our experiments. This resulted in consistent gains over the original dataset during the trial evaluation phase.

However during the final submission, in Subtask A we found that the model's performance in the test set did not correlate well with that of the validation set as shown in Table \ref{table:1}. This could be because the percentage of positive labels in the test set is only 10\% while the validation set has 50\%. 

Experiments without upsampling gives better performance in test set even though there is a decrease in the validation score. For Subtask B however, upsampling has actually increased the model performance. On hindsight, this could be because of similar distribution of class labels in both validation and test sets.

The submitted models received an $F_1$-score of 68.07 in Subtask A and 81.03 in Subtask B.

\subsection{Statistical Significance Test} \label{stat}
\citet{Reichart2018TheHG} suggests methods to measure whether two models have statistically significant differences in their predictions on a single dataset. We incorporate a non-parametric testing method for significance called the McNemar's test recommended by them for binary classification. Pairwise comparison of few of our models are reported in Table \ref{table:3}. The table contains the p-values for the null hypothesis. The null hypothesis is that two models do not have significant differences in their label predictions. In simpler words, a small p-value for an experiment pair denotes a significant difference in the prediction disagreement between two models. For example, from Table \ref{table:3}, DAN + GloVe and DAN + BERT models have a  p-value less than 0.05 in both sub-tasks. This indicates that there is significant disagreement between the predictions of two models. Since DAN + BERT gets a better F$_{1}$-score and p $<$ 0.05, we can confidently assert that improvement is not obtained by chance.

We use majority voting from five random seeds to get the final predictions on the test set for doing the paired significance testing.

\section{Conclusion} \label{conclusion}
We discussed our experiments for doing suggestion mining using tri-training. Tri-training combined with BERT representations proved to be an effective technique for doing semi-supervised learning especially in a cross-domain setting.
Future work could explore more optimal ways of doing tri-training, evaluate the effect of contextual representations in tri-training convergence, and try more sophisticated architectures for classification that may include different attention mechanisms.

\bibliography{zoho_semeval_2019}
\bibliographystyle{acl_natbib}

\end{document}